\documentclass{article}

\PassOptionsToPackage{numbers,compress}{natbib}

\usepackage[preprint]{neurips_2026}

\usepackage[utf8]{inputenc} 
\usepackage[T1]{fontenc}    
\usepackage{hyperref}       
\usepackage{url}            
\usepackage{booktabs}       
\usepackage{amsfonts}       
\usepackage{nicefrac}       
\usepackage{microtype}      
\usepackage{xcolor}         
\usepackage[pdftex]{graphicx}
\usepackage{booktabs, multirow}

\usepackage{xcolor}

\usepackage{rotating}

\usepackage{amsmath}

\title{Same Encoder, Different Winner: A Paired-View Framework for Cell Painting Encoder Evaluation}

\author{%
  \textbf{Tim Treis}$^{1,2,3}$ \quad
  \textbf{Nikita Moshkov}$^{1}$ \quad
  \textbf{Johan Fredin Haslum}$^{3}$ \quad \\ 
  \textbf{Shantanu Singh}$^{3}$ \quad
  \textbf{Fabian J. Theis}$^{1,2}$ \\[4pt]
  $^{1}$Institute of Computational Biology, Helmholtz Munich \\
  $^{2}$School of Computation, Information and Technology, Technical University of Munich \\
  $^{3}$Broad Institute of MIT and Harvard \\[4pt]
  \texttt{tim.treis@helmholtz-munich.de}
}

\begin{document}

\maketitle

\vspace*{-0.5cm}

\begin{abstract}
Vision encoders for Cell Painting are typically ranked by a single evaluation, commonly replicate mean average precision (mAP). We introduce CP-BG-Bench, a paired-view evaluation framework that holds the central cell fixed across four matched views (raw crop C, segmented S, and density-augmented variants CD and SD), ablating or augmenting surrounding pixels as a controlled intervention. Instantiating the framework on three datasets (JUMP-CP, RxRx1, RxRx3-core) and three encoder families (DINOv3 ViT-B/16, OpenPhenom, SubCell) under four community-standard protocols (replicate mAP, scIB batch integration, CellProfiler feature prediction, cross-batch perturbation recall), we find that the four protocols produce systematically different rankings of the same encoders, with disagreements decomposing along three orthogonal axes: cell versus background, morphology versus context, and within-study versus across-batch. The largest effect is between within-study replicate mAP and cross-batch recall: on RxRx3-core, SubCell with segmented inputs retains 94\% of crop replicate mAP (0.716 vs.\ 0.760) but only 32\% of crop R@10 (0.100 vs.\ 0.313), indicating that the within-study signal preserved under segmentation is largely non-transferable; density augmentation on the same configuration recovers 84\% of the within-study C-to-S gap but only 8\% of the cross-batch gap. Two further effects reinforce the decomposition: segmented views predict CellProfiler features as well as or better than crops on JUMP-CP and RxRx3-core, inverting the replicate-mAP ranking; and the C-to-S replicate-mAP gap varies by an order of magnitude across datasets (0.03 on JUMP-CP to 0.36 on RxRx1) with the magnitude similar across encoders within each dataset (with sign reversals only at the smallest gaps), indicating that the size of background-driven gain is set by experimental design rather than by the encoder. Single-metric ranking of Cell Painting encoders is therefore sensitive to which protocol is used, and the disagreements between protocols are interpretable as projections onto the three axes the paired-view design exposes. We will release the paired-view datasets, reconstruction pipelines, 36 trained checkpoints, aggregated embeddings, and the full evaluation suite.
\end{abstract}

\section{Introduction}

Image-based morphological profiling has become a central tool in phenotypic drug discovery and functional genomics, with Cell Painting~\cite{bray2016cellpainting,cimini2023optimizing} emerging as a widely used assay. Cell Painting stains six subcellular compartments with multiplexed fluorescent dyes, producing high-content images that summarize how perturbations alter cellular state. Until recently, profiles were constructed by segmenting cells and extracting hand-crafted features with tools such as CellProfiler~\cite{Carpenter2006-mh, stirling2021cellprofiler}, followed by per-plate normalization and aggregation~\cite{Caicedo2017-strategies}. Vision encoders trained directly on Cell Painting images now routinely match or exceed these classical features on standard benchmarks~\cite{Kraus2024-mk,Kim2025-nl,Moshkov2024-oj, SanchezFernandez2023-cloome}, and a growing ecosystem of foundation-style encoders is being developed against this evaluation backdrop.

The community has converged on a small toolkit of evaluation metrics. Replicate mean average precision (mAP), as implemented in the \texttt{copairs}~\cite{Kalinin2025-ok} library, measures whether replicate wells of the same perturbation cluster together across plates and is a commonly reported metric in published Cell Painting representation papers. The scIB~\cite{luecken2022benchmarking} suite, originally developed for single-cell omics, quantifies batch correction and biological conservation jointly. Cross-batch retrieval on a held-out experimental batch tests transfer to unseen acquisition conditions. New encoders are typically benchmarked on one or two of these and declared an improvement. What is not established is whether the metrics agree, or what each individually measures.

\begin{figure}[!h]
  \centering
  \includegraphics[width=\textwidth]{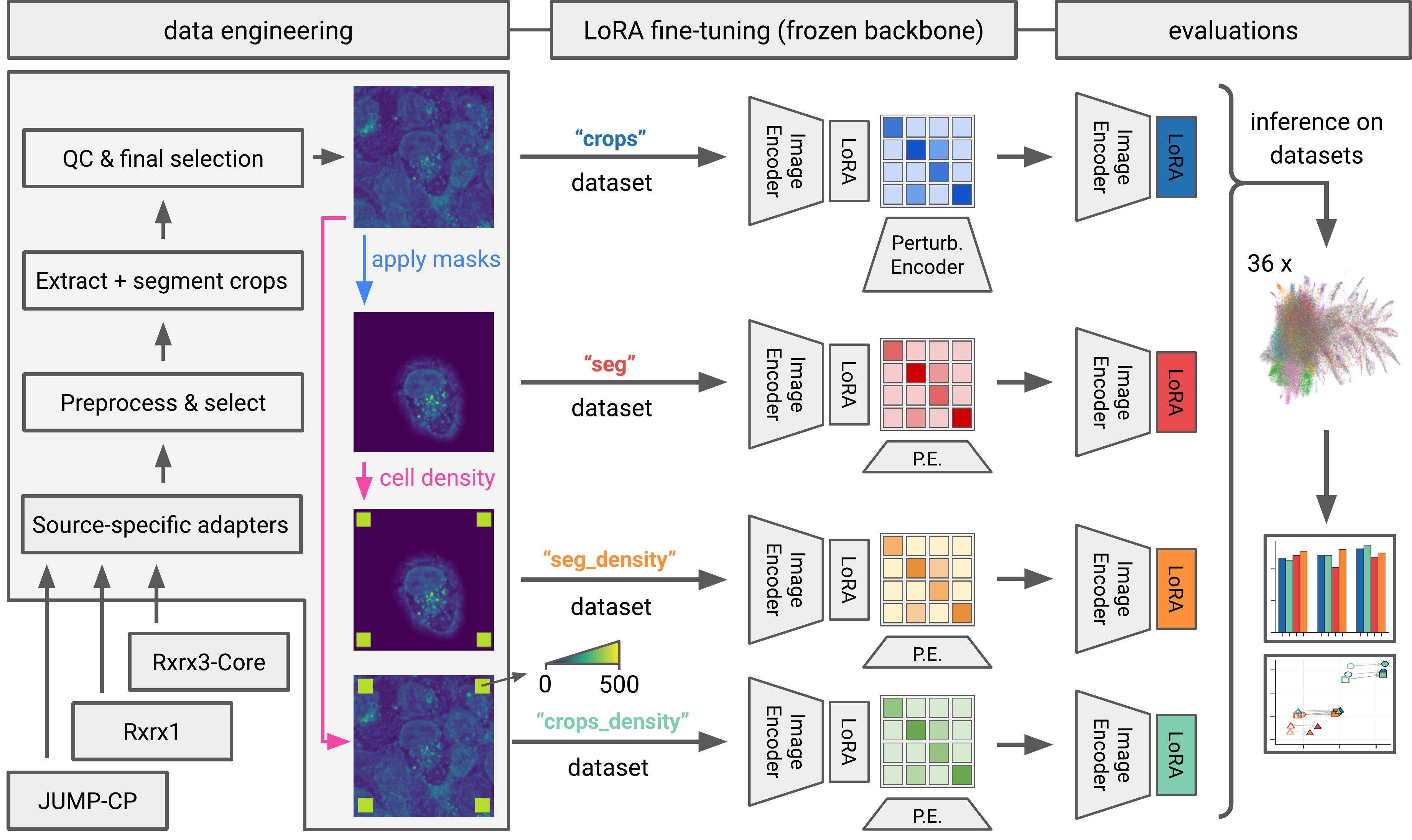}
  \caption{\label{fig:cp_bg_bench_overview}Overview of the CP-BG-Bench framework. \textbf{Left:} data engineering pipeline. Source-specific adapters ingest JUMP-CP, RxRx1, and RxRx3-core; images are preprocessed, single-cell crops are extracted, and per-cell quality control yields the final cell pool. \textbf{Center:} four matched views are derived from each cell tile while preserving identity. The raw crop (C); segmented (S) zeroes out everything outside the cell mask; crops with density patches (CD) and segmented with density patches (SD) overlay four corner patches whose intensity encodes local cell count, providing an explicit, source-agnostic density signal. \textbf{Right:} each of the twelve resulting encoder-view configurations per dataset (3 encoders $\times$ 4 views) is fine-tuned with LoRA against perturbation embeddings under a multi-positive contrastive objective, producing 36 trained checkpoints in total. The resulting embeddings are evaluated under four community-standard protocols (replicate mAP, scIB batch integration, CellProfiler feature prediction, cross-batch perturbation recall).}
\end{figure}

Two recent observations suggest the question is consequential. \citet{Seal2026} showed that cell count alone, computed without any image representation, predicts bioactivity benchmarks substantially above chance, indicating that within-study evaluations can be satisfied by signal that has no morphological content. \citet{Moshkov2024-oj} reported that representations trained on ``cells-in-context'' crops, which retain neighbouring cells and background pixels, outperform representations of tightly-cropped single cells on retrieval benchmarks; the authors attributed this to beneficial contextual cues, but the effect is also consistent with exploitation of within-batch artifacts that correlate with perturbation identity. Distinguishing these explanations requires a controlled comparison in which the central cell is held fixed while surrounding pixels are ablated or augmented. Existing Cell Painting datasets do not support such a comparison: crops always include background, and segmentation always removes it, leaving cell, neighbours, density, and background irrecoverably entangled. CP-BG-Bench (Figure~\ref{fig:cp_bg_bench_overview}) addresses this by providing every single-cell tile in four matched views that share the same cell centre, holding cellular morphology fixed while toggling background and density as controlled interventions.

\newpage
We address this with CP-BG-Bench (Figure~\ref{fig:cp_bg_bench_overview}), a paired-view evaluation framework in which every single-cell tile is provided in four matched views that share the same cell centre. The raw crop (C) and segmented cell (S) bracket the cell-versus-background contrast at the input level; the density-augmented variants (CD, SD) overlay an explicit, source-agnostic encoding of local cell density that is otherwise entangled with both. Holding the central cell fixed while toggling background context and density isolates each as a controlled intervention on the input. We instantiate the framework on three Cell Painting datasets spanning compounds (a JUMP-CP~\cite{Chandrasekaran2023-it} subset of cpg0016 in U2OS cells, 1{,}466 perturbations), siRNA knockdown (RxRx1~\cite{sypetkowski2023rxrx1datasetevaluatingexperimental} in HEPG2 cells, 1{,}108 perturbations), and CRISPR knockout (RxRx3-core~\cite{kraus2025rxrx3core} in HUVEC cells, 735 perturbations), and three vision encoders spanning different pretraining strategies (DINOv3 ViT-B/16~\cite{simeoni2025dinov3}, OpenPhenom~\cite{Kraus2024-mk}, SubCell~\cite{gupta2025subcell}). Each of the twelve resulting encoder-view configurations per dataset is fine-tuned with LoRA under a contrastive objective conditioned on perturbation embeddings (ECFP4 fingerprints for compounds~\cite{rogers2010ecfp4}, ESM2 embeddings for genetic perturbations~\cite{lin2023esm2}) and evaluated under four community-standard protocols: replicate mAP, scIB batch integration, linear prediction of CellProfiler features on a held-out batch, and cross-batch perturbation recall.

Across this matrix, the four protocols produce systematically different rankings of the same encoders, and the disagreements decompose along three orthogonal axes: cell versus background, morphology versus context, and within-study versus across-batch. The largest effect is on the within-versus-across-batch axis: SubCell on RxRx3-core retains 94\% of crop replicate mAP using segmented views (0.716 vs.\ 0.760) but only 32\% of crop R@10 (0.100 vs.\ 0.313), and density augmentation on the same configuration recovers 84\% of the within-study C-to-S gap but only 8\% of the cross-batch gap. On the morphology-versus-context axis, segmented views predict CellProfiler features as well as or better than crops on JUMP-CP and RxRx3-core, inverting the replicate-mAP ranking. On the cell-versus-background axis, the C-to-S replicate-mAP gap varies by an order of magnitude across datasets (0.03 on JUMP-CP to 0.36 on RxRx1) with the magnitude similar across encoders within each dataset (with sign reversals only at the smallest gaps), indicating that the size of background-driven gain is set by experimental design rather than by the encoder. Single-metric ranking of Cell Painting encoders is therefore sensitive to which protocol is used, and the disagreements between protocols are interpretable as projections onto the three axes the paired-view design exposes.

The contribution of this paper is methodological. CP-BG-Bench is the framework, the four-protocol evaluation on three datasets is its first deployment, and the systematic disagreement we observe is the empirical finding. We will release the paired-view JUMP-CP dataset under Creative Commons Zero v1.0, reconstruction pipelines that materialise the four-view structure for RxRx1 (CC BY-NC) and RxRx3-core from their respective public sources (we do not redistribute the source images, to comply with the originating licences), 36 trained encoder checkpoints, aggregated well-level embeddings, and the full evaluation suite. 

\section{Results}

We organise the results around the three axes the paired-view design exposes (cell versus background, morphology versus context, within-versus-across-batch); we report post-Harmony embeddings throughout, with Harmony's effect detailed in Appendix~\ref{app:batch_integration}.

\subsection{Background exploitability varies across datasets}
\label{sec:bg_exploitability}

\begin{figure}[!h]
  \centering
  \includegraphics[width=\textwidth]{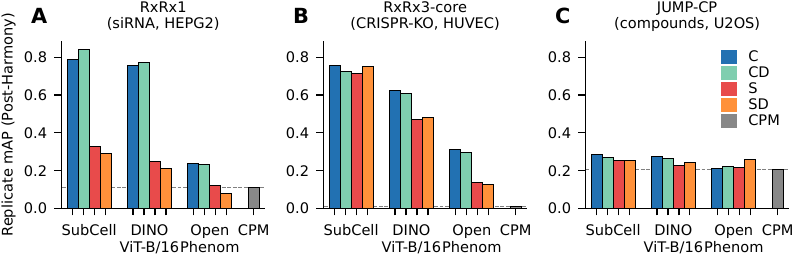}
  \caption{\label{fig:bg_exploitability_post}Replicate matching accuracy (mAP) across three datasets after Harmony batch correction. \textbf{(A)}~RxRx1 (siRNA, HEPG2), \textbf{(B)}~RxRx3-core (CRISPR-KO, HUVEC), \textbf{(C)}~JUMP-CP (compounds, U2OS). Bars show replicate mAP for the four views (C, CD, S, SD) extracted by three encoders (SubCell, DINO ViT-B/16, OpenPhenom). Grey bars show \texttt{cp\_measure} features as a baseline, processed through the same Spherize+PCA+Harmony pipeline. The crop-to-segmented gap is large on RxRx1, moderate on RxRx3-core, and minimal on JUMP-CP.}
\end{figure}

We measure this axis through the gap between crop-based views (C, CD), which retain background context, and segmented views (S, SD), which isolate cellular morphology, in replicate mAP across plates (Figure~\ref{fig:bg_exploitability_post}). Replicate mAP, computed with the \texttt{copairs} library~\cite{Kalinin2025-ok}, scores how reliably wells of the same perturbation cluster together across plates, treating same-perturbation wells on different plates as positives and treated-control pairs on the same plate as negatives; we report the macro-averaged value over treated wells of post-Harmony, well-level embeddings, with the encoders LoRA-adapted under a multi-positive contrastive objective conditioned on perturbation identity (Methods Sections~\ref{sec:finetuning},~\ref{sec:aggregation},~\ref{sec:evaluation}).

The C-to-S gap varies substantially across datasets. On RxRx1, crop views reach replicate mAP of 0.76--0.84 (SubCell, DINO) and 0.24 (OpenPhenom), while segmented views drop to 0.08--0.33; the gap exceeds 0.4 for the two stronger encoders. RxRx3-core shows a similar but attenuated pattern, with SubCell retaining most of its crop performance under segmentation while DINO and OpenPhenom lose substantial signal. JUMP-CP, in contrast, shows uniformly low replicate mAP across all views and encoders (0.21--0.29), and a near-absent C-to-S gap. The magnitude of the gap is similar across the three encoders within each dataset (with OpenPhenom on JUMP-CP showing a marginal sign reversal: S=0.214 vs.\ C=0.209), indicating that the size of background-driven gain is set by experimental design rather than by the encoder.

We use CellProfiler features, extracted via \texttt{cp\_measure}~\cite{munoz2025cpmeasure}, as a baseline (CPM) that ranks the three datasets in approximately the opposite order to the finetuned encoders. The three datasets show three different patterns. RxRx1 and RxRx3-core are produced by a single imaging facility under tightly-controlled acquisition conditions, so background variation is constrained and any residual variation can become correlated with perturbation identity through plate layout, cell density, and neighbour effects, inflating within-study replicate mAP for crop-based views. On JUMP-CP, two factors compound to suppress crop-driven gain: at the data-generation stage, randomised plate layouts across 10 imaging sources~\cite{Arevalo2024-kv} for the compound plates decorrelate background from perturbation identity; at the training stage, our cross-batch-stratified minibatch sampler (Section~\ref{sec:finetuning}) encourages the contrastive objective to favour features invariant to source-driven background variation, since same-perturbation positives within a minibatch span multiple sources. The C-to-S gap therefore reflects how well background variation is structured around perturbation identity under the training procedure used. That CPM collapses to 0.011 on RxRx3-core, where learned segmented views reach 0.75, indicates that classical features after our matched preprocessing do not separate CRISPR-KO HUVEC replicates. The comparison is between unsupervised features (CPM) and supervised representations (learned encoders fine-tuned against perturbation identity); the gap reflects supervision rather than purely architectural advantage.

To validate that batch correction via Harmony isn't the driving force behind the results shown in Section~\ref{sec:bg_exploitability}, we evaluate all encoder-view configurations with the scIB benchmark suite~\cite{luecken2022benchmarking} before and after integration. This jointly quantifies batch correction and biological conservation in a single aggregate "Total score" (Table~\ref{tab:scib_summary}).

\begin{table}[h]
\centering
\caption{Mean Harmony delta ($\Delta$ = post $-$ pre) across all 12 encoder-view configurations per dataset. Harmony improves batch correction substantially while leaving biological conservation unchanged.}
\label{tab:scib_summary}
\small
\begin{tabular}{lccc}
\toprule
Dataset & $\Delta$ Total & $\Delta$ Batch & $\Delta$ Bio \\
\midrule
RxRx1      & +0.103 & +0.244 & +0.009  \\
RxRx3-core & +0.074 & +0.185 & +0.000  \\
JUMP-CP    & +0.070 & +0.186 & $-$0.007 \\
\bottomrule
\end{tabular}
\end{table}

Harmony consistently improves integration across all three datasets, with "Total score" gains of +0.070 to +0.103. This improvement is mostly driven by batch correction ($\Delta$ +0.185 to +0.244), while biological conservation remains essentially unchanged ($\Delta$ ranging from $-$0.007 to +0.009). The slight negative $\Delta$ Bio on JUMP-CP is consistent with its combination of strong batch effects and weak perturbation signal (Section~\ref{sec:bg_exploitability}), which makes batch and biological variance harder to disentangle. Crucially, the near-zero biological conservation deltas across all datasets indicate that the post-Harmony retrieval results reported throughout this work reflect genuine batch correction rather than artificial inflation of biological similarity ( Appendix~\ref{app:batch_integration}).

\subsection{Background signal does not transfer across batches}
\label{sec:perturbation_recall}

The replicate mAP evaluation in Section~\ref{sec:bg_exploitability} operates within the full embedding space, where query and gallery wells share the same imaging batches. To test whether the signal that distinguishes views transfers to unseen batches, we evaluate cross-batch perturbation recall: queries from a single held-out batch must retrieve same-perturbation wells from the remaining batches. Results are shown in Table~\ref{tab:recall}.

\begin{table}[!h]
\centering
\caption{Cross-batch perturbation recall (R@1 / R@10, post-Harmony) on the held-out
batch. Higher values indicate better cross-batch generalization. Bold indicates best
R@10 per dataset.}
\label{tab:recall}
\small
\begin{tabular}{llcccc}
\toprule
 & & \multicolumn{4}{c}{R@1 / R@10} \\
\cmidrule(lr){3-6}
Dataset & Encoder & C & CD & S & SD \\
\midrule
\multirow{3}{*}{RxRx1}
  & SubCell    & 0.474 / 0.676 & 0.467 / 0.653 & 0.192 / 0.409 & 0.181 / 0.369 \\
  & DINO       & 0.552 / \textbf{0.742} & 0.519 / 0.716 & 0.137 / 0.322 & 0.154 / 0.368 \\
  & OpenPhenom & 0.285 / 0.568 & 0.264 / 0.564 & 0.071 / 0.258 & 0.042 / 0.162 \\
\midrule
\multirow{3}{*}{RxRx3-core}
  & SubCell    & 0.086 / \textbf{0.313} & 0.079 / 0.300 & 0.019 / 0.100 & 0.023 / 0.117 \\
  & DINO       & 0.075 / 0.296 & 0.071 / 0.302 & 0.025 / 0.123 & 0.023 / 0.131 \\
  & OpenPhenom & 0.020 / 0.121 & 0.015 / 0.093 & 0.004 / 0.035 & 0.004 / 0.033 \\
\midrule
\multirow{3}{*}{JUMP-CP}
  & SubCell    & 0.057 / \textbf{0.170} & 0.060 / 0.163 & 0.045 / 0.140 & 0.059 / 0.150 \\
  & DINO       & 0.054 / 0.156 & 0.051 / 0.158 & 0.029 / 0.108 & 0.034 / 0.130 \\
  & OpenPhenom & 0.022 / 0.094 & 0.018 / 0.101 & 0.010 / 0.050 & 0.014 / 0.059 \\
\bottomrule
\end{tabular}
\end{table}

The C-to-S gap on cross-batch recall is proportionally far larger than on replicate mAP. On RxRx3-core, SubCell~S retains 94\% of C's replicate mAP (0.716 vs.\ 0.760) but only 32\% of its R@10 (0.100 vs.\ 0.313). RxRx1 is starker still: DINO~C achieves R@10 of 0.742 while DINO~S drops to 0.322. On JUMP-CP, where the cell-versus-background gap is small to begin with, the recall gap is correspondingly small (SubCell C: 0.170 vs.\ S: 0.140). To quantify the dependence directly: on RxRx1, segmented views recover only 33--51\% of crop replicate mAP and 43--60\% of crop R@10 depending on the encoder; the remainder is driven by pixels outside the cell. On JUMP-CP, segmented views recover 82--102\% of crop replicate mAP, (i.e. S matches or marginally exceeds C for OpenPhenom), consistent with background contributing little on multi-site datasets. Background signal is batch-specific: it correlates with perturbation identity within a study but does not transfer when queries and gallery come from different batches.

Density augmentation, which recovers 84\% of the within-study C-to-S gap on RxRx3-core (Section~\ref{sec:bg_exploitability}), recovers only 8\% of the cross-batch gap (SD R@10 0.117 vs.\ S 0.100, against C 0.313). On RxRx1, SD actually underperforms S for SubCell (0.369 vs.\ 0.409) and OpenPhenom (0.162 vs.\ 0.258). Adding density to crop views (CD) does not help either, underperforming C by up to 0.03 in cross-batch R@10 on RxRx1 (0.004 for OpenPhenom, 0.023--0.027 for SubCell and DINO).

\subsection{Morphology versus context: the background signal is not classical morphology}
\label{sec:cp_prediction}

The preceding sections establish that background inflates within-study replicate mAP and that the inflating signal poorly transfers across batches. This raises the question of whether the learnt embedding captures morphology that interpretable hand-crafted features would also capture. We use linear predictability of CellProfiler features (computed via \texttt{cp\_measure} on the same wells) as an operational proxy for morphology recapitulation: we fit a multi-output ridge regression from each embedding and evaluate on a held-out batch (Methods Section~\ref{sec:evaluation}).

\begin{figure}[!h]
  \centering
  \includegraphics[width=\textwidth]{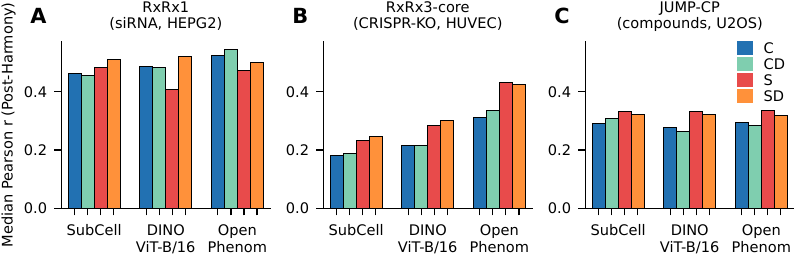}
  \caption{\label{fig:cpm_pred}Median Pearson $r$ between embedding-predicted and observed \texttt{cp\_measure} features on the held-out batch (post-Harmony). Each panel shows a different dataset: \textbf{(A)}~RxRx1 (siRNA, HEPG2), \textbf{(B)}~RxRx3-core (CRISPR-KO, HUVEC), \textbf{(C)}~JUMP-CP (compounds, U2OS). Bars indicate median Pearson $r$ across 1,358--1,443 \texttt{cp\_measure} features for four views (C, CD, S, SD) extracted by three encoders (SubCell, DINO ViT-B/16, OpenPhenom). On JUMP-CP and RxRx3-core, segmented views (S, SD) consistently outperform crop views (C, CD), inverting the replicate mAP ranking (Figure~\ref{fig:bg_exploitability_post}). On RxRx1, the pattern is encoder-dependent: SD matches or exceeds both crop variants for SubCell and DINO but falls below both crops for OpenPhenom. The background signal that inflates replicate mAP does not correspond to classical morphological content.}
\end{figure}

In Figure~\ref{fig:cpm_pred} we can see that the ranking inverts: On JUMP-CP and RxRx3-core, segmented views predict \texttt{cp\_measure} features better than crop views despite achieving comparable or lower replicate mAP (Section~\ref{sec:bg_exploitability}). The inversion is sharpest on RxRx3-core: OpenPhenom~S achieves median $r$ of 0.432 compared to 0.311 for OpenPhenom~C. On JUMP-CP, segmented views consistently outperform crops by 0.02--0.06 in median $r$.

The scores for RxRx1 present as more nuanced (Table~\ref{tab:cp_prediction}): S alone underperforms C for DINO (0.409 vs.\ 0.487) and OpenPhenom (0.474 vs.\ 0.526), but SD recovers and exceeds crop performance for SubCell (0.511 vs.\ max(C,CD)=0.464) and DINO (0.521 vs.\ 0.487), while OpenPhenom shows the opposite pattern (SD 0.500 below CD 0.546). The best segmented configuration on RxRx1 (DINO SD, 0.521) outperforms four of the six crop configurations (all SubCell and DINO crops) and trails OpenPhenom CD (0.546) by 0.025, despite crops dominating replicate mAP by a factor of two to three (Figure~\ref{fig:bg_exploitability_post}). The background signal that inflates replicate mAP therefore does not systematically improve prediction of these cell-only features.

\begin{table}[!h]
\centering
\caption{Median Pearson $r$ between embedding-predicted and observed \texttt{cp\_measure} features on the held-out batch (post-Harmony). Higher values indicate greater linear accessibility of classical morphological information. Bold indicates best per dataset.}
\label{tab:cp_prediction}
\small
\begin{tabular}{llcccc}
\toprule
Dataset & Encoder & C & CD & S & SD \\
\midrule
\multirow{3}{*}{RxRx1}
  & SubCell         & 0.464 & 0.455 & 0.485 & 0.511 \\
  & DINO            & 0.487 & 0.484 & 0.409 & 0.521 \\
  & OpenPhenom      & 0.526 & \textbf{0.546} & 0.474 & 0.500 \\
\midrule
\multirow{3}{*}{RxRx3-core}
  & SubCell         & 0.183 & 0.189 & 0.233 & 0.247 \\
  & DINO            & 0.217 & 0.215 & 0.283 & 0.302 \\
  & OpenPhenom      & 0.311 & 0.335 & \textbf{0.432} & 0.425 \\
\midrule
\multirow{3}{*}{JUMP-CP}
  & SubCell         & 0.290 & 0.307 & 0.332 & 0.324 \\
  & DINO            & 0.278 & 0.263 & 0.332 & 0.322 \\
  & OpenPhenom      & 0.293 & 0.283 & \textbf{0.335} & 0.319 \\
\bottomrule
\end{tabular}
\end{table}

Harmony has dataset-dependent effects on CP prediction scores (Appendix~\ref{app:cp_prediction}): negligible on RxRx3-core, modest on JUMP-CP, and a meaningful lift on RxRx1 (especially for DINO). The qualitative ranking of views is unchanged across pre- and post-Harmony in all cases.

A per-feature-group breakdown (Appendix~\ref{app:cp_prediction}) reveals a consistent predictability hierarchy across all encoders, views, and datasets: intensity and texture features are the best predicted (median $r$ 0.28--0.68; 0.40--0.68 on RxRx1 and JUMP-CP), while shape features (AreaShape, Zernike) are poorly recovered (0.02--0.34). Whether this reflects encoder architecture, pretraining distribution, or the contrastive objective rewarding whichever features best separate perturbations is left to the Discussion.

\section{Methods}

\subsection{Datasets}
\label{sec:datasets}

We instantiate CP-BG-Bench on three Cell Painting datasets spanning different perturbation modalities and experimental designs (Table~\ref{tab:datasets}).

\begin{table}[h]
\centering
\caption{Dataset summary. Well counts include controls.}
\label{tab:datasets}
\small
\begin{tabular}{lllccc}
\toprule
Dataset & Perturbation type & Cell line & Wells & Perturbations & Batches \\
\midrule
JUMP-CP    & Compounds     & U2OS  & 28,003  & 1,466 & 10 sources \\
RxRx1      & siRNA         & HEPG2 & 12,581  & 1,108 & 11 sources \\
RxRx3-core & CRISPR-KO     & HUVEC & 104,166 & 735   & 9 plates \\
\bottomrule
\end{tabular}
\end{table}

Each dataset designates one batch as the held-out test set (JUMP-CP: source\_4; RxRx1: HEPG2-07; RxRx3-core: plate\_4). The held-out set is used for early stopping and cross-batch evals, all remaining batches are used for training and within-study evaluation. We will release the paired-view JUMP-CP dataset under Creative Commons Zero v1.0, with Croissant metadata including Responsible AI fields. To comply with the licence terms of RxRx1 (CC BY-NC) and RxRx3-core, we provide a deterministic reconstruction pipeline that materialises the four-view structure from each dataset's official public release. All artifacts (datasets, reconstruction pipelines, trained checkpoints, aggregated embeddings, and the evaluation suite) will be released publicly with the final version of this manuscript.\footnote{Release links will be added in a future revision of this preprint.}

\subsection{Paired-view construction}
\label{sec:view_construction}

For each field of view, we segment nuclei and cells using Cellpose-SAM~\cite{pachitariu2025cellposesam} and extract a crop around each cell centroid. From each crop (C), three additional views are derived while preserving cell identity:
\textbf{Segmented (S):} all pixels outside the cell and nucleus masks are zeroed.
\textbf{Crop + Density (CD):} four corner patches are drawn on the crop, with intensity encoding local cell density (FOV cell count, min-max scaled).
\textbf{Segmented + Density (SD):} the same density patches applied to the segmented view.
The density patch provides an explicit, source-agnostic encoding of local cell crowding, isolating density-driven signal from both background context and cellular morphology. A schema can be found in Figure~\ref{fig:cp_bg_bench_overview}, full construction details (crop dimensions, interpolation, QC filters) are in Appendix~\ref{app:view_construction}.

\subsection{Encoders and fine-tuning}
\label{sec:finetuning}

We evaluate three vision encoders: \textbf{DINO ViT-B/16}~\cite{simeoni2025dinov3}, pretrained on natural images (LVD-1689M) with a trainable channel adapter for Cell Painting input; \textbf{OpenPhenom}~\cite{Kraus2024-mk}, pretrained with masked autoencoding on RxRx3~\cite{fay2023rxrx3}, natively accepting multi-channel fluorescence; and \textbf{SubCell}~\cite{gupta2025subcell}, pretrained with masked autoencoding on Human Protein Atlas images with a trainable channel adapter. All encoders are adapted with LoRA~\cite{hu2021loralowrankadaptationlarge} (frozen backbone) and project into a shared 128-dimensional embedding space.

Encoders are trained with a symmetric multi-positive contrastive loss~\cite{khosla2020supervised} that aligns image embeddings with perturbation embeddings: ECFP4 fingerprints for compounds (JUMP-CP) and ESM2 embeddings for genetic perturbations (RxRx1, RxRx3-core). Each minibatch contains 512 samples drawn from 64 perturbation classes, stratified across imaging batches to ensure cross-batch coverage within each perturbation. This stratification matters for interpretation: same-perturbation positives within a minibatch span multiple imaging batches, so the contrastive objective receives gradient signal that favours features invariant to batch-specific background variation (Section~\ref{sec:bg_exploitability}). Architecture details, hyperparameters, and the full training protocol are in Appendix~\ref{app:training}.

\subsection{Aggregation and batch correction}
\label{sec:aggregation}

Single-cell embeddings are L2-normalised, mean-pooled per well, spherised (ZCA-cor whitening fit on controls), reduced to 50 PCA components, and optionally Harmony-corrected~\cite{korsunsky2019harmony}. Both pre- and post-Harmony embeddings are retained for all downstream evaluations. Details are in Appendix~\ref{app:aggregation}.

\subsection{Baselines}
\label{sec:baselines}

We compute classical morphological features per cell using \texttt{cp\_measure} (intensity, texture, granularity, shape, and correlation modules across all channels and compartments), yielding 1,358--1,443 features per dataset post feature selection as described in \texttt{pycytominer}~\cite{Serrano2025-sn}. To enable direct comparison with learned embeddings, \texttt{cp\_measure} features undergo the same aggregation and batch-correction pipeline. The baseline participates in replicate mAP evaluation and serves as the prediction target for the CP feature prediction probe. Details are in Appendix~\ref{app:baselines}.

\subsection{Evaluation protocols}
\label{sec:evaluation}

All evaluations are conducted on well-level embeddings in both pre- and post-Harmony spaces. Control wells are excluded except for batch integration.

\paragraph{Replicate mAP.} We use \texttt{copairs}~\cite{Kalinin2025-ok} to compute per-perturbation average precision over cross-plate replicate retrieval. Positives are wells sharing the same perturbation on different plates; negatives are treated-control pairs on the same plate.

\paragraph{Batch integration.} We evaluate with the scIB suite~\cite{luecken2022benchmarking}, reporting batch correction, biological conservation, and a weighted Total score ($0.4 \times \text{Batch} + 0.6 \times \text{Bio}$).

\paragraph{CP feature prediction.} We fit a multi-output ridge regression from the 50-d embedding to all \texttt{cp\_measure} features, trained on non-held-out wells and evaluated on the held-out batch. We report median Pearson $r$ across features, which is offset-invariant within the test batch.

\paragraph{Perturbation recall.} Non-control wells in the held-out batch serve as queries, all non-control wells in remaining batches form the gallery. We report macro-averaged R@$k$ ($k{=}1{\ldots}10$) by cosine similarity.

\section{Discussion}

For practitioners ranking encoders today, the implication is direct: single-metric rankings should be treated as tentative. The C-to-S replicate-mAP gap is a useful diagnostic; a large gap signals that a substantial fraction of an encoder's reported score may not transfer to new acquisition conditions, and a configuration with lower replicate mAP but higher CP-prediction or cross-batch recall is the better choice when generalisation matters. Concretely, on RxRx1 segmented views recover only 33--51\% of crop replicate mAP depending on encoder; the remainder reflects signal that does not survive a change of acquisition batch. For dataset designers, the results reframe JUMP-CP. Its multi-site protocol, combined with our cross-batch-stratified training procedure, suppresses the correlation between background and perturbation identity. The low absolute mAP in JUMP-CP, pointing to it being a difficult dataset, is consistent with a design that suppresses the shortcut single-facility datasets inadvertently reward. 

The four protocols project onto the three axes differently, and this is what makes the disagreements informative rather than noisy. Replicate mAP rewards within-study clustering of perturbation labels and is largely indifferent to whether the clustering signal lives in cellular morphology or in surrounding context, which is why it inflates on datasets where background context correlates with perturbation identity. CellProfiler-feature prediction is offset-invariant within the test batch and isolates whether the embedding linearly encodes classical photometric and textural morphology; its agreement with replicate mAP on JUMP-CP and disagreement on RxRx1 and RxRx3-core localises the signal in each case. Cross-batch perturbation recall depends on absolute embedding geometry across acquisition conditions and penalises any signal that does not transfer. This suggests, it is the most conservative of the four and the most relevant when deployment involves data from unseen plates or sites. Within-study clustering remains the appropriate metric when the use case is mechanism-of-action discovery within a single screen. The scIB suite tracks closely with replicate mAP because biological conservation is dominated by similar within-study geometry, making it a less independent check than its separate provenance might suggest. An encoder's score on any single metric is a partial summary; the four scores presented in this study together expose distinct aspects of what the encoder has captured.

\newpage
Table~\ref{tab:metric_claims} summarises what each protocol measures, the assumptions under which it operates, and the failure mode the paired-view design exposes.

\begin{table}[h]
\centering
\caption{What each evaluation protocol measures and where it fails. The paired-view design exposes the failure mode in the rightmost column.}
\label{tab:metric_claims}
\small
\begin{tabular}{p{1.7cm}p{3.2cm}p{3.5cm}p{4.0cm}}
\toprule
Protocol & What it measures & Key assumption & Failure mode exposed \\
\midrule
Replicate mAP & Within-study clustering of perturbation replicates across plates & Signal that clusters replicates is biological & Inflated by background context correlated with perturbation identity (C-to-S gap up to 0.51) \\
\midrule
scIB Total & Joint batch correction and biological conservation & Bio-conservation score reflects perturbation biology & Tracks replicate mAP; dominated by within-study geometry, inherits its failure modes \\
\midrule
CP feature\newline prediction & Linear accessibility of classical morphological features in the embedding & CellProfiler features are a sufficient proxy for interpretable morphology & Segmented views outperform crops (ranking inverts), showing the extra signal in crops is not classical morphology \\
\midrule
Cross-batch recall & Perturbation retrieval across unseen acquisition conditions & Embedding geometry is stable across batches & Crop advantage collapses (94\% mAP retained, 32\% R@10 retained); density recovery drops from 84\% to 8\% \\
\bottomrule
\end{tabular}
\end{table}

The framework has bounded scope. We evaluate three Cell Painting datasets, three encoder families, a single fine-tuning recipe (LoRA under a multi-positive contrastive objective with cross-batch-stratified sampling), a single batch correction method (Harmony), and a single segmentation pipeline (Cellpose-SAM). The three datasets intentionally differ simultaneously in perturbation modality, cell line, replication depth, and batch structure, preventing attribution of the different C-to-S gap magnitudes to any single factor. The cross-dataset and cross-view contrasts on which our central claims rest are large in absolute terms (C-to-S replicate-mAP gaps of 0.03 to 0.51; cross-batch R@10 gaps of 0.03 to 0.42 between C and S) and exceed typical training-seed variance reported in similar contrastive fine-tuning settings, but we report a single fine-tuning run per configuration and a multi-seed replication would strengthen confidence in the smaller within-dataset effects, in particular the CP-prediction inversions on RxRx1 where the C-to-S gap is on the order of 0.05. The cross-batch-stratified minibatch sampler interacts with dataset design: on multi-site datasets like JUMP-CP, same-perturbation positives within a minibatch span multiple sources, providing gradient signal against background-correlated features; single-facility datasets lack the source diversity within minibatches needed for the same effect. Alternative samplers or objectives may produce different gap magnitudes on the same data.

Two open questions follow. First, whether the cross-batch transfer gap can be reduced by training objectives that explicitly penalise within-study-only signal, for example by including cross-batch positives during contrastive training or by adversarial regularisation against batch-correlated context. Second, whether the universal predictability hierarchy we observe (intensity and texture features well predicted, AreaShape and Zernike poorly recovered) reflects an architectural limitation of ViT-based encoders, a property of fluorescence-image pretraining distributions, or a consequence of the contrastive objective, since geometric features may not be discriminative for perturbation identity at the single-cell scale even when they are biologically meaningful. CP-BG-Bench is designed to make these questions answerable by providing the paired views and standardised evaluation protocols under which competing training objectives or architectures can be compared on equal footing.

\newpage
{
  \small
  \bibliographystyle{unsrtnat}
  \bibliography{references}
}


\appendix
\renewcommand{\thefigure}{A\arabic{figure}}
\setcounter{figure}{0}
\renewcommand{\thetable}{A\arabic{table}}
\setcounter{table}{0}

\newpage
\section{Appendix}

\subsection{View Construction Details}
\label{app:view_construction}

For each field of view (FOV), we segment nuclei and cells using Cellpose-SAM with default parameters and dataset-specific channel routing (DNA for nuclei; AGP+DNA or Phalloidin+DNA for cells, depending on dataset). Nucleus-cell pairs are matched by mask overlap; unmatched objects are discarded. A 150$\times$150 pixel crop is extracted around each cell centroid, excluding cells whose centroid falls within 75 pixels of the FOV border. Crops are resized to 224$\times$224 using bilinear interpolation for fluorescent channels and nearest-neighbour interpolation for mask channels.

Quality filtering removes border cells and cells outside the $[0.025, 0.975]$ quantile range on nuclear area, cytoplasmic area, and nucleus-to-cytoplasm ratio. Cells are sampled via \texttt{uniform\_per\_compound\_source} with a target of 100 cells per perturbation-source combination (JUMP-CP, RxRx1) or 200 (RxRx3-core); per-well cell counts vary widely.

For density-augmented views (CD, SD), four solid square patches (22$\times$22 pixels, inset 7 pixels from each corner) are drawn on every channel. Patch intensity encodes local cell density: the number of cells in the originating FOV, min-max scaled to $[0, 255]$ across the cell pool.

For segmented views (S, SD), the DNA channel is zeroed outside the nucleus mask; all other fluorescent channels are zeroed outside the cell mask.

\subsection{Encoder Architecture and Training Details}
\label{app:training}

\paragraph{Encoder architectures.}
All encoders are adapted with Low-Rank Adaptation (LoRA; $r{=}8$, $\alpha{=}16$, dropout 0.1) applied to all attention and MLP projections in every transformer block, with the remaining trunk parameters frozen. Each encoder feeds into a shared projection head: a two-layer MLP with GeGLU activation and skip connection, mapping to a 128-dimensional embedding.

\textbf{DINO ViT-B/16:} A ViT-B/16 backbone pretrained with DINOv3 on LVD-1689M (\texttt{timm/vit\_base\_patch16\_dinov3.lvd1689m}), yielding 768-dimensional CLS tokens. Cell Painting images (5 or 6 fluorescent channels) are mapped to the 3-channel input via a trainable 1$\times$1 convolution (no bias), followed by a learned affine normalisation initialised to ImageNet statistics.

\textbf{OpenPhenom:} A channel-aware ViT (\texttt{recursionpharma/OpenPhenom}) pretrained with masked autoencoding~\cite{He2021MaskedLearners} on RxRx3. The model natively accepts 5-channel input at 256$\times$256, producing 384-dimensional embeddings, requiring no channel adapter.

\textbf{SubCell:} A ViT-B/16 backbone pretrained with masked autoencoding on Human Protein Atlas images, with a learned two-head attention pooling layer producing 1536-dimensional embeddings. SubCell expects 4 channels (microtubules, ER, DNA, protein); the remaining Cell Painting channels are combined into the protein slot via a trainable 1$\times$1 convolution. Input images are min-max normalised per image and resized to 224$\times$224.

\paragraph{Contrastive loss.}
Image and perturbation embeddings are independently L2-normalised, and logits are computed as $\ell_{ij} = \tau^{-1} \cdot \mathbf{z}_i^{\text{img}} \cdot {\mathbf{z}_j^{\text{pert}}}^\top$ where $\tau$ is a learnable temperature initialised to 0.1 and clamped to $[0.01, 100]$. For each anchor $i$ with positive set $P(i) = \{j : \text{label}_j = \text{label}_i\}$, the loss minimises the negative log of the total softmax mass placed on label-equal partners:
\begin{equation}
\mathcal{L}_i = -\log \sum_{j \in P(i)}
\frac{\exp(\ell_{ij})}{\sum_{k} \exp(\ell_{ik})}
\end{equation}
applied symmetrically in image-to-perturbation and perturbation-to-image directions with equal weight. This is the in-log multi-positive variant of contrastive learning~\cite{khosla2020supervised}, distinct from standard CLIP~\cite{radford2021clip} (one positive per anchor) and SigLIP~\cite{zhai2023siglip} (pairwise sigmoid).

\paragraph{Perturbation embeddings.}
For compound screens (JUMP-CP), perturbations are represented by ECFP4 fingerprints (Morgan radius 2, 2048 bits, chirality-aware, log$(1{+}\text{count})$-transformed), precomputed per InChIKey. For genetic perturbation screens (RxRx1, RxRx3-core), perturbations are represented by ESM2 embeddings (\texttt{facebook/esm2\_t33\_650M\_UR50D}, mean-pooled over sequence tokens, 1280-dimensional), precomputed per gene symbol. Control wells (DMSO, EMPTY) receive the mean embedding across all perturbations. Perturbation features are passed through a trainable projection head (two-layer MLP with SiLU activation and skip connection) mapping to the shared 128-dimensional space. No LoRA is applied to the perturbation branch.

\paragraph{Training protocol.}
Models are trained for up to 25 epochs with early stopping (patience 5, monitoring macro-averaged image-to-perturbation R@1 on a held-out validation set). Optimisation uses AdamW ($\beta_1{=}0.9$, $\beta_2{=}0.95$) with a cosine learning rate schedule and 10\% linear warmup. Learning rates are 3$\times$10$^{-4}$ for projection heads, 1$\times$10$^{-4}$ for LoRA parameters, and 1$\times$10$^{-5}$ for the temperature scalar. Weight decay of 5$\times$10$^{-4}$ is applied to projection heads only. Training uses bf16 mixed precision with gradient clipping at 5.

Each batch contains 512 samples drawn from 64 perturbation classes (8 samples per class), with samples stratified across imaging batches to ensure cross-plate coverage within each perturbation. Control compounds are pinned into every batch. Augmentations consist of random horizontal and vertical flips, random 0--90\textdegree{} rotation, and light brightness/contrast jitter ($\pm$10\%).

\subsection{Aggregation and Batch Correction Details}
\label{app:aggregation}

Single-cell embeddings are L2-normalised immediately after the encoder, then mean-pooled per well. Well-level embeddings are spherised (ZCA-cor whitening fit on control wells only, applied to all wells) using pycytominer, followed by PCA reduction to 50 components.

Batch correction is performed with harmonypy using default parameters ($\theta{=}2$, $\lambda{=}1$, $\sigma{=}0.1$, max iterations 100) on the 50-dimensional PCA embeddings. The batch key is imaging source for RxRx1 (11 sources) and JUMP-CP (10 sources), and plate for RxRx3-core (9 plates). Both pre-Harmony (\texttt{X\_pca}) and post-Harmony (\texttt{X\_pca\_harmony}) embeddings are retained for all downstream evaluations. No re-normalisation is applied after Harmony.

\subsection{Baseline and Evaluation Protocol Details}
\label{app:baselines}

\paragraph{\texttt{cp\_measure} features.}
Classical morphological features are computed per cell using \texttt{cp\_measure} across five per-channel modules (intensity, texture, granularity, radial distribution, radial Zernikes) applied to each channel under both nucleus and cytoplasm masks, three shape modules (size/shape, Zernike, Feret) per compartment, and three pairwise correlation measures (Pearson, Manders fold, RWC) for all channel pairs. Per-cell features are mean-aggregated to well level, low-coverage features (NaN fraction $>$ 50\%) and zero-variance-in-controls features are dropped, then well-level features are spherised (ZCA-cor whitening fit on control wells only), reduced to 50 PCA components, and Harmony-corrected with the same parameters as the learned-embedding pipeline. This yields 1,358--1,443 features per dataset entering PCA.

\paragraph{Replicate mAP.}
Per-perturbation mAP is obtained by averaging AP values across treated wells only, with statistical significance assessed by a permutation null (10,000 permutations, $\alpha{=}0.05$, Benjamini-Hochberg corrected). Treated-treated pairs on the same plate are excluded via the copairs reference-index mechanism.

\paragraph{Batch integration.}
Batch correction is quantified by iLISI, kBET, BRAS, graph connectivity, and PCR comparison (5 metrics). Biological conservation is quantified by cLISI, KMeans NMI, KMeans ARI, silhouette label, and isolated labels (5 metrics). Each component score is the unweighted mean of its constituent metrics. The batch key is imaging source (JUMP-CP, RxRx1) or plate (RxRx3-core); the label key is perturbation identity (InChIKey for compounds, gene symbol for genetic perturbations).

\paragraph{CP feature prediction.}
Ridge regression uses $\alpha{=}1.0$. Embeddings are standardised (zero mean, unit variance) with statistics fit on the training set only, then applied to both train and test. The model is fit on all non-control, non-held-out wells and evaluated on non-control wells in the held-out batch. Features with near-zero variance on the training set ($<$10$^{-8}$) are excluded.

\newpage
\subsection{Extended Background Exploitability Analysis}

\begin{figure}[!h]
  \centering
  \includegraphics[width=\textwidth]{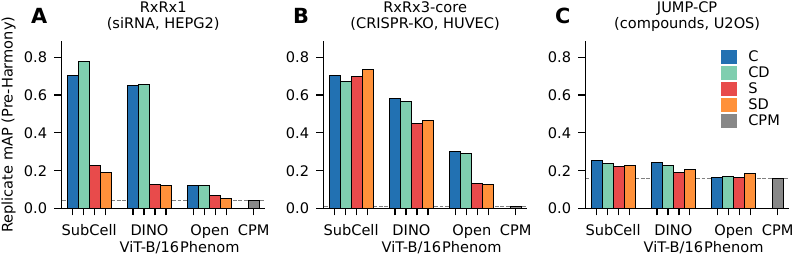}
  \caption{\label{fig:bg_exploitability_pre}Replicate matching accuracy (mAP) across three datasets before Harmony batch correction. Each panel shows a different dataset and perturbation modality: (A) RxRx1 (siRNA, HEPG2), (B) RxRx3-core (CRISPR-KO, HUVEC), (C) JUMP-CP (compounds, U2OS). Bars indicate replicate mAP for the four views of the data (crops C, crops\_density CD, seg S, and seg\_density SD) extracted by three vision encoders (SubCell, DINO ViT-B/16, OpenPhenom). Grey bars show the mAP of reference features extracted by \texttt{cp\_measure} as a baseline. Higher values indicate stronger replicate reproducibility. The performance gap between crop-based views (C, CD) and segmented views (S, SD) quantifies background exploitability and varies substantially across datasets, indicating that background signal contribution is a dataset-specific property.}
\end{figure}

\begin{table}[!h]
\centering
\caption{Replicate matching accuracy (mAP) across three datasets, before and after Harmony batch correction. Rows show three vision encoders (SubCell, DINO ViT-B/16, OpenPhenom) and a classical \texttt{cp\_measure} (CPM) baseline. Columns show four views of the data: crops (C), crops with density context (CD), segmented (S), and segmented with density context (SD). Bold indicates the best value per dataset within each Harmony stage. The CPM baseline operates on segmented cells only; other view columns are not applicable.}
\label{tab:map_replicate}
\small
\begin{tabular}{llcccccccc}
\toprule
 & & \multicolumn{4}{c}{Pre-Harmony} & \multicolumn{4}{c}{Post-Harmony} \\
\cmidrule(lr){3-6} \cmidrule(lr){7-10}
Dataset & Encoder & C & CD & S & SD & C & CD & S & SD \\
\midrule
\multirow{4}{*}{RxRx1}
  & SubCell    & 0.705 & \textbf{0.780} & 0.225 & 0.192 & 0.787 & \textbf{0.843} & 0.330 & 0.290 \\
  & DINO       & 0.649 & 0.655          & 0.124 & 0.121 & 0.756 & 0.773          & 0.251 & 0.209 \\
  & OpenPhenom & 0.121 & 0.119          & 0.070 & 0.050 & 0.239 & 0.233          & 0.122 & 0.077 \\
  & CPM        & -     & -              & 0.039 & - & -     & -              & 0.108 & - \\
\midrule
\multirow{4}{*}{RxRx3-core}
  & SubCell    & 0.705 & 0.674 & 0.702 & \textbf{0.739} & \textbf{0.760} & 0.725 & 0.716 & 0.753 \\
  & DINO       & 0.582 & 0.569 & 0.452 & 0.466          & 0.626          & 0.611 & 0.469 & 0.484 \\
  & OpenPhenom & 0.303 & 0.289 & 0.133 & 0.125          & 0.312          & 0.296 & 0.136 & 0.127 \\
  & CPM        & -     & -     & 0.011 & -        & -              & -     & 0.011 & - \\
\midrule
\multirow{4}{*}{JUMP-CP}
  & SubCell    & \textbf{0.255} & 0.240 & 0.220 & 0.229 & \textbf{0.285} & 0.270 & 0.252 & 0.256 \\
  & DINO       & 0.241          & 0.227 & 0.192 & 0.205 & 0.276          & 0.263 & 0.225 & 0.243 \\
  & OpenPhenom & 0.161          & 0.166 & 0.166 & 0.185 & 0.209          & 0.222 & 0.214 & 0.257 \\
  & CPM        & -              & -     & 0.156 & - & -            & -     & 0.205 & - \\
\bottomrule
\end{tabular}
\end{table}

\newpage
\subsection{Extended Batch Integration Analysis}
\label{app:batch_integration}

This section provides the full per-configuration scIB benchmark results. Table~\ref{tab:scib_summary} summarises the mean Harmony delta across all 12 encoder-view configurations per dataset. We evaluate batch integration quality using the scIB suite~\cite{luecken2022benchmarking}, reporting three composite scores: Total (weighted aggregate), Batch Correction (mean of iLISI, kBET, BRAS, graph connectivity), and Bio Conservation (mean of cLISI, NMI, ARI, silhouette label, isolated labels).

\begin{figure}[!h]
\centering
\includegraphics[width=\textwidth]{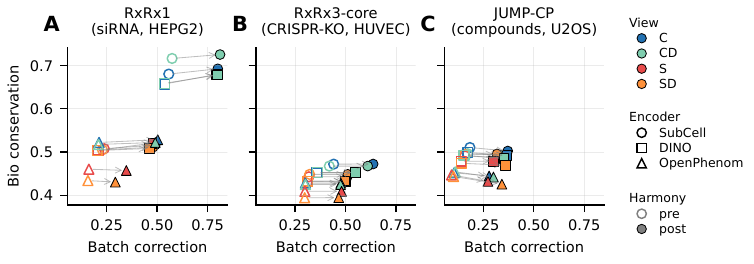}
\caption{Batch correction vs.\ biological conservation (scIB) for all encoder-view configurations before (open markers) and after (filled markers) Harmony. Each panel shows a different dataset: \textbf{(A)}~RxRx1, \textbf{(B)}~RxRx3-core, \textbf{(C)}~JUMP-CP. Marker color indicates the encoder; marker shape indicates the view. Arrows connect matched pre/post configurations. All configurations improve in batch correction with negligible change in biological conservation.}
\label{fig:batch_scatter}
\end{figure}

Figure~\ref{fig:batch_scatter} visualizes the effect of Harmony on each encoder-view configuration. Every configuration shifts rightward (improved batch correction) without moving downward (preserved biological conservation), confirming that Harmony operates as intended across all encoders, views, and datasets.

Two dataset-specific patterns are worth noting. On RxRx1, crop-based views (C, CD) achieve substantially higher biological conservation than segmented views (S, SD) both before and after Harmony (e.g., SubCell CD: 0.726 vs.\ SubCell S: 0.516 post-Harmony). This gap confirms that background context contributes to biological discriminability on this dataset, consistent with the replicate matching results in Section~\ref{sec:bg_exploitability}. On JUMP-CP, biological conservation shows a slight negative trend under Harmony ($\Delta$ Bio = $-$0.007 on average, worst case: OpenPhenom SD at $-$0.018), suggesting minimal overcorrection. This is consistent with JUMP-CP's combination of strong batch effects and weak perturbation signal (replicate mAP 0.21-0.29; Section~\ref{sec:bg_exploitability}), which makes batch and biological variance harder to disentangle, leaving Harmony more likely to remove small amounts of biological structure alongside batch.

Additionally, OpenPhenom segmented views on RxRx1 are outliers in terms of Harmony effectiveness: these are the worst-performing configurations in replicate matching (mAP 0.122 and 0.077; Section~\ref{sec:bg_exploitability}, Table~\ref{tab:map_replicate}), indicating that when the embedding contains very little biological signal, Harmony has less meaningful structure to correct.

Tables~\ref{tab:scib_pre_full} and~\ref{tab:scib_post_full} provide the complete per-configuration scores before and after Harmony.

\clearpage
\begin{sidewaystable}[!h]
\centering
\caption{Full scIB scores (individual metrics), pre-Harmony. Batch = mean(iLISI, kBET, BRAS, Graph conn.), Bio = mean(cLISI, NMI, ARI, Silhouette, Isolated). Bold indicates best Total per dataset. PCR comparison is 0.000 for all configurations pre-Harmony by construction (the pre-integration embedding serves as the reference) and is omitted.}
\label{tab:scib_pre_full}
\footnotesize
\begin{tabular}{llccccccccccccc}
\toprule
 & & \multicolumn{4}{c}{Batch correction} & \multicolumn{5}{c}{Bio conservation} & & & \\
\cmidrule(lr){3-6} \cmidrule(lr){7-11}
Dataset & Encoder / View & iLISI & kBET & BRAS & Graph & cLISI & NMI & ARI & Silh. & Isol. & Batch & Bio & Total \\
\midrule
\multirow{12}{*}{RxRx1}
  & SubCell C      & 0.514 & 0.703 & 0.699 & 0.872 & 0.997 & 0.849 & 0.420 & 0.552 & 0.584 & 0.558 & 0.680 & 0.631 \\
  & SubCell CD     & 0.562 & 0.693 & 0.696 & 0.919 & 0.998 & 0.890 & 0.548 & 0.567 & 0.579 & 0.574 & 0.716 & \textbf{0.659} \\
  & SubCell S      & 0.180 & 0.101 & 0.667 & 0.243 & 0.991 & 0.643 & 0.032 & 0.442 & 0.429 & 0.238 & 0.507 & 0.400 \\
  & SubCell SD     & 0.169 & 0.072 & 0.683 & 0.222 & 0.991 & 0.641 & 0.027 & 0.440 & 0.426 & 0.229 & 0.505 & 0.395 \\
  & DINO C         & 0.491 & 0.722 & 0.642 & 0.836 & 0.997 & 0.827 & 0.358 & 0.545 & 0.560 & 0.538 & 0.657 & 0.610 \\
  & DINO CD        & 0.475 & 0.719 & 0.649 & 0.839 & 0.997 & 0.833 & 0.370 & 0.547 & 0.547 & 0.536 & 0.659 & 0.610 \\
  & DINO S         & 0.091 & 0.049 & 0.744 & 0.164 & 0.991 & 0.648 & 0.042 & 0.428 & 0.411 & 0.210 & 0.504 & 0.386 \\
  & DINO SD        & 0.123 & 0.026 & 0.687 & 0.183 & 0.990 & 0.643 & 0.018 & 0.437 & 0.429 & 0.204 & 0.503 & 0.384 \\
  & OpenPhenom C   & 0.135 & 0.038 & 0.646 & 0.242 & 0.990 & 0.661 & 0.025 & 0.458 & 0.476 & 0.212 & 0.522 & 0.398 \\
  & OpenPhenom CD  & 0.146 & 0.034 & 0.638 & 0.229 & 0.990 & 0.661 & 0.022 & 0.453 & 0.461 & 0.209 & 0.517 & 0.394 \\
  & OpenPhenom S   & 0.182 & 0.002 & 0.496 & 0.130 & 0.989 & 0.599 & 0.005 & 0.368 & 0.341 & 0.162 & 0.460 & 0.341 \\
  & OpenPhenom SD  & 0.226 & 0.000 & 0.447 & 0.114 & 0.988 & 0.550 & 0.001 & 0.332 & 0.297 & 0.157 & 0.434 & 0.323 \\
\midrule
\multirow{12}{*}{RxRx3-core}
  & SubCell C      & 0.538 & 0.623 & 0.923 & 0.124 & 0.991 & 0.416 & 0.005 & 0.449 & 0.498 & 0.442 & 0.472 & \textbf{0.460} \\
  & SubCell CD     & 0.539 & 0.523 & 0.923 & 0.115 & 0.991 & 0.404 & 0.005 & 0.446 & 0.491 & 0.420 & 0.467 & 0.448 \\
  & SubCell S      & 0.566 & 0.045 & 0.942 & 0.041 & 0.992 & 0.311 & 0.002 & 0.429 & 0.478 & 0.319 & 0.443 & 0.393 \\
  & SubCell SD     & 0.567 & 0.057 & 0.939 & 0.050 & 0.993 & 0.324 & 0.002 & 0.434 & 0.488 & 0.323 & 0.448 & 0.398 \\
  & DINO C         & 0.534 & 0.238 & 0.928 & 0.083 & 0.990 & 0.373 & 0.004 & 0.436 & 0.458 & 0.357 & 0.452 & 0.414 \\
  & DINO CD        & 0.533 & 0.276 & 0.929 & 0.070 & 0.990 & 0.373 & 0.004 & 0.434 & 0.460 & 0.362 & 0.452 & 0.416 \\
  & DINO S         & 0.558 & 0.024 & 0.946 & 0.027 & 0.991 & 0.288 & 0.002 & 0.429 & 0.449 & 0.311 & 0.432 & 0.383 \\
  & DINO SD        & 0.557 & 0.037 & 0.943 & 0.031 & 0.991 & 0.296 & 0.003 & 0.426 & 0.447 & 0.314 & 0.432 & 0.385 \\
  & OpenPhenom C   & 0.553 & 0.017 & 0.910 & 0.029 & 0.989 & 0.282 & 0.002 & 0.416 & 0.452 & 0.302 & 0.428 & 0.377 \\
  & OpenPhenom CD  & 0.567 & 0.011 & 0.907 & 0.026 & 0.989 & 0.271 & 0.002 & 0.411 & 0.450 & 0.302 & 0.425 & 0.376 \\
  & OpenPhenom S   & 0.580 & 0.009 & 0.887 & 0.019 & 0.989 & 0.246 & 0.001 & 0.391 & 0.418 & 0.299 & 0.409 & 0.365 \\
  & OpenPhenom SD  & 0.587 & 0.009 & 0.876 & 0.018 & 0.988 & 0.234 & 0.002 & 0.366 & 0.381 & 0.298 & 0.394 & 0.356 \\
\midrule
\multirow{12}{*}{JUMP-CP}
  & SubCell C      & 0.000 & 0.030 & 0.685 & 0.204 & 0.991 & 0.564 & 0.021 & 0.428 & 0.548 & 0.184 & 0.510 & \textbf{0.380} \\
  & SubCell CD     & 0.000 & 0.021 & 0.656 & 0.186 & 0.991 & 0.543 & 0.011 & 0.418 & 0.516 & 0.173 & 0.496 & 0.367 \\
  & SubCell S      & 0.000 & 0.019 & 0.609 & 0.165 & 0.991 & 0.543 & 0.012 & 0.401 & 0.509 & 0.159 & 0.491 & 0.358 \\
  & SubCell SD     & 0.000 & 0.020 & 0.593 & 0.170 & 0.991 & 0.539 & 0.014 & 0.405 & 0.524 & 0.157 & 0.494 & 0.359 \\
  & DINO C         & 0.000 & 0.031 & 0.649 & 0.187 & 0.991 & 0.546 & 0.011 & 0.415 & 0.532 & 0.174 & 0.499 & 0.369 \\
  & DINO CD        & 0.000 & 0.018 & 0.584 & 0.171 & 0.991 & 0.528 & 0.008 & 0.407 & 0.527 & 0.155 & 0.492 & 0.357 \\
  & DINO S         & 0.000 & 0.009 & 0.560 & 0.132 & 0.991 & 0.520 & 0.005 & 0.400 & 0.477 & 0.140 & 0.478 & 0.343 \\
  & DINO SD        & 0.000 & 0.013 & 0.556 & 0.143 & 0.991 & 0.517 & 0.005 & 0.394 & 0.462 & 0.142 & 0.474 & 0.341 \\
  & OpenPhenom C   & 0.000 & 0.003 & 0.414 & 0.113 & 0.991 & 0.507 & 0.003 & 0.372 & 0.386 & 0.106 & 0.452 & 0.314 \\
  & OpenPhenom CD  & 0.000 & 0.002 & 0.441 & 0.112 & 0.991 & 0.504 & 0.003 & 0.374 & 0.398 & 0.111 & 0.454 & 0.317 \\
  & OpenPhenom S   & 0.000 & 0.003 & 0.351 & 0.110 & 0.991 & 0.505 & 0.002 & 0.326 & 0.410 & 0.093 & 0.447 & 0.305 \\
  & OpenPhenom SD  & 0.001 & 0.003 & 0.394 & 0.114 & 0.991 & 0.499 & 0.002 & 0.327 & 0.397 & 0.102 & 0.443 & 0.307 \\
\bottomrule
\end{tabular}
\end{sidewaystable}

\newpage
\clearpage
\begin{sidewaystable}
\centering
\caption{Full scIB scores (individual metrics), post-Harmony. Batch = mean(iLISI, kBET, BRAS, Graph conn.), Bio = mean(cLISI, NMI, ARI, Silhouette, Isolated). Bold indicates best Total per dataset.}
\label{tab:scib_post_full}
\footnotesize
\begin{tabular}{llccccccccccccc}
\toprule
 & & \multicolumn{4}{c}{Batch correction} & \multicolumn{5}{c}{Bio conservation} & & & \\
\cmidrule(lr){3-6} \cmidrule(lr){7-11}
Dataset & Encoder / View & iLISI & kBET & BRAS & Graph & cLISI & NMI & ARI & Silh. & Isol. & Batch & Bio & Total \\
\midrule
\multirow{12}{*}{RxRx1}
  & SubCell C      & 0.583 & 0.822 & 0.743 & 0.894 & 0.998 & 0.865 & 0.456 & 0.560 & 0.583 & 0.800 & 0.693 & 0.736 \\
  & SubCell CD     & 0.616 & 0.825 & 0.742 & 0.937 & 0.998 & 0.900 & 0.574 & 0.577 & 0.579 & 0.812 & 0.726 & \textbf{0.760} \\
  & SubCell S      & 0.361 & 0.132 & 0.730 & 0.263 & 0.991 & 0.648 & 0.039 & 0.457 & 0.444 & 0.491 & 0.516 & 0.506 \\
  & SubCell SD     & 0.362 & 0.097 & 0.743 & 0.245 & 0.991 & 0.644 & 0.037 & 0.452 & 0.431 & 0.480 & 0.511 & 0.499 \\
  & DINO C         & 0.575 & 0.840 & 0.711 & 0.876 & 0.998 & 0.855 & 0.421 & 0.556 & 0.568 & 0.795 & 0.680 & 0.726 \\
  & DINO CD        & 0.572 & 0.850 & 0.720 & 0.879 & 0.998 & 0.859 & 0.422 & 0.560 & 0.554 & 0.799 & 0.679 & 0.727 \\
  & DINO S         & 0.276 & 0.164 & 0.788 & 0.215 & 0.991 & 0.659 & 0.050 & 0.460 & 0.444 & 0.480 & 0.521 & 0.504 \\
  & DINO SD        & 0.340 & 0.070 & 0.752 & 0.205 & 0.991 & 0.642 & 0.025 & 0.447 & 0.437 & 0.464 & 0.508 & 0.490 \\
  & OpenPhenom C   & 0.423 & 0.097 & 0.734 & 0.319 & 0.991 & 0.674 & 0.047 & 0.460 & 0.470 & 0.503 & 0.528 & 0.518 \\
  & OpenPhenom CD  & 0.419 & 0.085 & 0.724 & 0.305 & 0.991 & 0.674 & 0.043 & 0.457 & 0.446 & 0.491 & 0.522 & 0.509 \\
  & OpenPhenom S   & 0.383 & 0.005 & 0.582 & 0.151 & 0.989 & 0.600 & 0.008 & 0.364 & 0.325 & 0.347 & 0.457 & 0.413 \\
  & OpenPhenom SD  & 0.369 & 0.000 & 0.496 & 0.124 & 0.989 & 0.556 & 0.003 & 0.323 & 0.281 & 0.293 & 0.430 & 0.375 \\
\midrule
\multirow{12}{*}{RxRx3-core}
  & SubCell C      & 0.558 & 0.648 & 0.928 & 0.125 & 0.991 & 0.417 & 0.005 & 0.449 & 0.499 & 0.637 & 0.472 & \textbf{0.538} \\
  & SubCell CD     & 0.559 & 0.534 & 0.927 & 0.115 & 0.991 & 0.405 & 0.005 & 0.445 & 0.491 & 0.610 & 0.467 & 0.525 \\
  & SubCell S      & 0.572 & 0.044 & 0.943 & 0.040 & 0.992 & 0.312 & 0.002 & 0.429 & 0.479 & 0.509 & 0.443 & 0.469 \\
  & SubCell SD     & 0.573 & 0.057 & 0.940 & 0.050 & 0.992 & 0.324 & 0.002 & 0.434 & 0.489 & 0.512 & 0.448 & 0.474 \\
  & DINO C         & 0.557 & 0.245 & 0.931 & 0.081 & 0.990 & 0.374 & 0.004 & 0.435 & 0.459 & 0.548 & 0.452 & 0.491 \\
  & DINO CD        & 0.556 & 0.285 & 0.932 & 0.070 & 0.990 & 0.374 & 0.004 & 0.433 & 0.461 & 0.551 & 0.452 & 0.492 \\
  & DINO S         & 0.568 & 0.024 & 0.947 & 0.026 & 0.991 & 0.288 & 0.002 & 0.428 & 0.450 & 0.499 & 0.432 & 0.459 \\
  & DINO SD        & 0.567 & 0.037 & 0.944 & 0.031 & 0.991 & 0.296 & 0.003 & 0.425 & 0.450 & 0.497 & 0.433 & 0.459 \\
  & OpenPhenom C   & 0.564 & 0.018 & 0.912 & 0.029 & 0.989 & 0.282 & 0.002 & 0.416 & 0.452 & 0.477 & 0.428 & 0.448 \\
  & OpenPhenom CD  & 0.577 & 0.012 & 0.907 & 0.026 & 0.989 & 0.272 & 0.002 & 0.411 & 0.451 & 0.476 & 0.425 & 0.445 \\
  & OpenPhenom S   & 0.586 & 0.009 & 0.888 & 0.019 & 0.989 & 0.246 & 0.001 & 0.391 & 0.418 & 0.481 & 0.409 & 0.438 \\
  & OpenPhenom SD  & 0.591 & 0.009 & 0.877 & 0.018 & 0.988 & 0.234 & 0.002 & 0.366 & 0.382 & 0.467 & 0.394 & 0.423 \\
\midrule
\multirow{12}{*}{JUMP-CP}
  & SubCell C      & 0.027 & 0.067 & 0.777 & 0.204 & 0.992 & 0.570 & 0.027 & 0.428 & 0.494 & 0.371 & 0.502 & \textbf{0.450} \\
  & SubCell CD     & 0.020 & 0.063 & 0.739 & 0.188 & 0.992 & 0.550 & 0.016 & 0.413 & 0.493 & 0.345 & 0.493 & 0.434 \\
  & SubCell S      & 0.017 & 0.047 & 0.704 & 0.173 & 0.992 & 0.545 & 0.018 & 0.396 & 0.482 & 0.323 & 0.487 & 0.421 \\
  & SubCell SD     & 0.011 & 0.049 & 0.703 & 0.176 & 0.992 & 0.541 & 0.018 & 0.405 & 0.521 & 0.318 & 0.495 & 0.424 \\
  & DINO C         & 0.019 & 0.067 & 0.759 & 0.189 & 0.992 & 0.555 & 0.018 & 0.419 & 0.475 & 0.367 & 0.492 & 0.442 \\
  & DINO CD        & 0.013 & 0.059 & 0.725 & 0.177 & 0.992 & 0.531 & 0.012 & 0.415 & 0.477 & 0.353 & 0.485 & 0.432 \\
  & DINO S         & 0.007 & 0.038 & 0.675 & 0.145 & 0.992 & 0.525 & 0.007 & 0.394 & 0.477 & 0.302 & 0.479 & 0.408 \\
  & DINO SD        & 0.037 & 0.041 & 0.727 & 0.151 & 0.992 & 0.512 & 0.007 & 0.391 & 0.445 & 0.359 & 0.469 & 0.425 \\
  & OpenPhenom C   & 0.024 & 0.021 & 0.598 & 0.121 & 0.992 & 0.511 & 0.003 & 0.365 & 0.352 & 0.279 & 0.444 & 0.378 \\
  & OpenPhenom CD  & 0.037 & 0.023 & 0.654 & 0.123 & 0.992 & 0.503 & 0.003 & 0.352 & 0.360 & 0.300 & 0.442 & 0.385 \\
  & OpenPhenom S   & 0.115 & 0.012 & 0.527 & 0.111 & 0.992 & 0.495 & 0.003 & 0.321 & 0.350 & 0.273 & 0.432 & 0.368 \\
  & OpenPhenom SD  & 0.155 & 0.006 & 0.637 & 0.115 & 0.992 & 0.481 & 0.003 & 0.297 & 0.355 & 0.342 & 0.426 & 0.392 \\
\bottomrule
\end{tabular}
\end{sidewaystable}

\clearpage
\subsection{Extended CellProfiler Feature Prediction}
\label{app:cp_prediction}

\begin{figure}[!h]
\centering
\includegraphics[width=\textwidth]{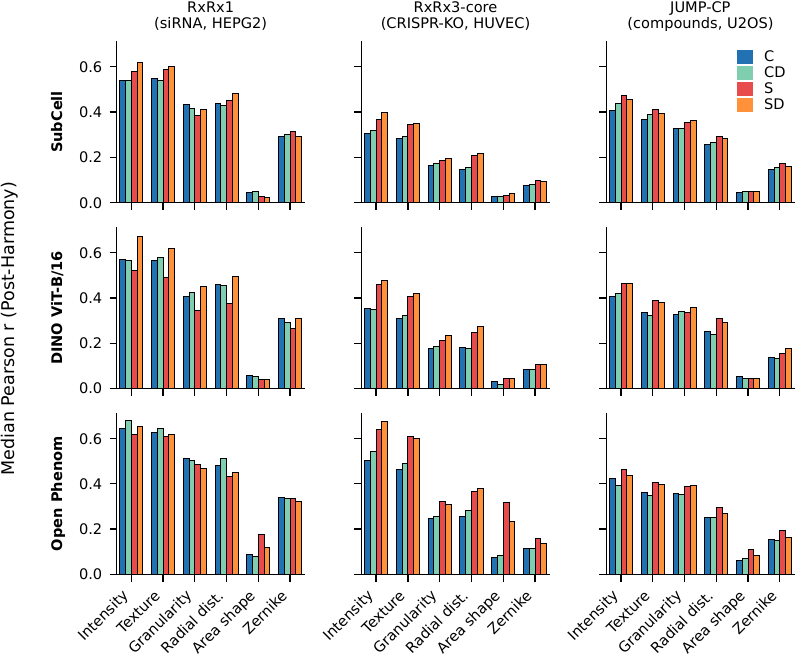}
\caption{Per-feature-group median Pearson $r$ for CP feature prediction (post-Harmony).
Rows correspond to encoders (SubCell, DINO ViT-B/16, OpenPhenom); columns to datasets
(\textbf{left}: RxRx1, \textbf{center}: RxRx3-core, \textbf{right}: JUMP-CP). Within
each panel, bars show the four views (C, CD, S, SD). The predictability hierarchy is
universal: intensity and texture features are well predicted across all configurations,
while shape features (AreaShape, Zernike) are poorly recovered. Segmented views (S, SD)
outperform crop views (C, CD) within most feature groups, particularly on RxRx3-core.}
\label{fig:cp_prediction_groups}
\end{figure}

This section provides the full per-configuration CellProfiler feature prediction results summarized in Section~\ref{sec:cp_prediction}. We report median Pearson $r$ between ridge-predicted and observed \texttt{cp\_measure} features on the held-out batch, for all encoder-view configurations in both pre- and post-Harmony embedding spaces (Table~\ref{tab:cp_prediction_full}), and a per-feature-group breakdown (Figure~\ref{fig:cp_prediction_groups}).

\begin{table}[!h]
\centering
\caption{Median Pearson $r$ for CP feature prediction across all encoder-view configurations, before and after Harmony. Bold indicates best per dataset within each Harmony stage.}
\label{tab:cp_prediction_full}
\small
\begin{tabular}{llcccccccc}
\toprule
 & & \multicolumn{4}{c}{Pre-Harmony} & \multicolumn{4}{c}{Post-Harmony} \\
\cmidrule(lr){3-6} \cmidrule(lr){7-10}
Dataset & Encoder & C & CD & S & SD & C & CD & S & SD \\
\midrule
\multirow{4}{*}{RxRx1}
  & SubCell         & 0.450 & 0.444 & 0.465 & 0.494 & 0.464 & 0.455 & 0.485 & 0.511 \\
  & DINO            & 0.451 & 0.447 & 0.355 & 0.487 & 0.487 & 0.484 & 0.409 & 0.521 \\
  & OpenPhenom      & 0.513 & \textbf{0.539} & 0.466 & 0.494 & 0.526 & \textbf{0.546} & 0.474 & 0.500 \\
\midrule
\multirow{4}{*}{RxRx3-core}
  & SubCell         & 0.181 & 0.189 & 0.232 & 0.246 & 0.183 & 0.189 & 0.233 & 0.247 \\
  & DINO            & 0.217 & 0.216 & 0.283 & 0.302 & 0.217 & 0.215 & 0.283 & 0.302 \\
  & OpenPhenom      & 0.311 & 0.334 & \textbf{0.433} & 0.425 & 0.311 & 0.335 & \textbf{0.432} & 0.425 \\
\midrule
\multirow{4}{*}{JUMP-CP}
  & SubCell         & 0.269 & 0.295 & 0.320 & 0.324 & 0.290 & 0.307 & 0.332 & 0.324 \\
  & DINO            & 0.274 & 0.256 & 0.299 & 0.315 & 0.278 & 0.263 & 0.332 & 0.322 \\
  & OpenPhenom      & 0.294 & 0.289 & \textbf{0.341} & 0.324 & 0.293 & 0.283 & \textbf{0.335} & 0.319 \\
\bottomrule
\end{tabular}
\end{table}

Two patterns emerge from the extended results. First, Harmony has a dataset-specific effect on CP prediction. On RxRx3-core, pre- and post-Harmony values differ by less than 0.002 across all configurations, confirming that what Harmony removes on this dataset is entirely orthogonal to classical morphology. On JUMP-CP, the per-config Harmony lift ranges from $-$0.006 to +0.033 (median +0.007), with the largest gains on SubCell~C (+0.021) and DINO~S (+0.033) while OpenPhenom configurations show negligible or slightly negative deltas. On RxRx1, the lift is encoder-dependent: DINO gains +0.034 to +0.054, SubCell +0.011 to +0.020, and OpenPhenom +0.006 to +0.013.

Second, Figure~\ref{fig:cp_prediction_groups} reveals a consistent predictability hierarchy across all datasets and encoders: intensity and texture features are the best predicted (median r 0.28–0.68 across configurations; 0.40–0.68 on RxRx1 / JUMP-CP), granularity and radial distribution occupy an intermediate range (0.15--0.51), while AreaShape and Zernike features are poorly recovered (0.02--0.34). This hierarchy is consistent with the embeddings encoding intensity and texture more accessibly than shape under our linear probe; whether shape information is genuinely absent or simply nonlinearly encoded cannot be determined from a ridge probe alone.

\clearpage
\subsection{Extended Perturbation Recall}
\label{app:perturbation_recall}

This section provides pre-Harmony recall results and the effect of Harmony on
cross-batch retrieval, complementing the post-Harmony results in
Section~\ref{sec:perturbation_recall}.

\begin{table}[!h]
\centering
\caption{Cross-batch perturbation recall (R@1 / R@10) before and after Harmony. Bold
indicates best R@10 per dataset within each Harmony stage.}
\label{tab:recall_full}
\small
\begin{tabular}{llcccc}
\toprule
 & & \multicolumn{4}{c}{Pre-Harmony R@1 / R@10} \\
\cmidrule(lr){3-6}
Dataset & Encoder & C & CD & S & SD \\
\midrule
\multirow{3}{*}{RxRx1}
  & SubCell    & 0.390 / 0.595 & 0.386 / 0.569 & 0.157 / 0.353 & 0.135 / 0.340 \\
  & DINO       & 0.492 / \textbf{0.699} & 0.464 / 0.684 & 0.110 / 0.276 & 0.138 / 0.351 \\
  & OpenPhenom & 0.222 / 0.491 & 0.170 / 0.438 & 0.049 / 0.203 & 0.027 / 0.128 \\
\midrule
\multirow{3}{*}{RxRx3-core}
  & SubCell    & 0.085 / \textbf{0.313} & 0.080 / 0.301 & 0.020 / 0.101 & 0.023 / 0.117 \\
  & DINO       & 0.075 / 0.297 & 0.072 / 0.305 & 0.023 / 0.123 & 0.024 / 0.132 \\
  & OpenPhenom & 0.019 / 0.118 & 0.014 / 0.093 & 0.004 / 0.035 & 0.004 / 0.035 \\
\midrule
\multirow{3}{*}{JUMP-CP}
  & SubCell    & 0.055 / \textbf{0.142} & 0.044 / 0.138 & 0.037 / 0.123 & 0.042 / 0.129 \\
  & DINO       & 0.047 / 0.121 & 0.053 / 0.148 & 0.026 / 0.077 & 0.033 / 0.101 \\
  & OpenPhenom & 0.019 / 0.101 & 0.021 / 0.106 & 0.011 / 0.047 & 0.017 / 0.060 \\
\bottomrule
\end{tabular}
\end{table}

Harmony has a strongly dataset-dependent effect on cross-batch recall. On RxRx3-core, pre- and post-Harmony recall values are identical to within 0.003 across all configurations, mirroring the no-op behavior observed for CP feature prediction (Appendix~\ref{app:cp_prediction}). On JUMP-CP, Harmony provides a modest lift in R@10 (+0.01 to +0.035 for SubCell and DINO, negligible for OpenPhenom).

On RxRx1, by contrast, Harmony substantially improves recall: crop-based views gain +0.03 to +0.13 in R@10, with the largest effect on OpenPhenom~CD (+0.094 R@1, +0.126 R@10). Segmented views also benefit (+0.02 to +0.06 R@10 for 5 of 6 configurations; DINO SD is +0.017) though the absolute gains are smaller. This is the opposite pattern from CP feature prediction, where RxRx1 showed the smallest Harmony effect (Appendix~\ref{app:cp_prediction}). The difference is consistent with the two metrics measuring different aspects of the embedding: Pearson~$r$ is offset-invariant and therefore insensitive to the batch-level shifts that Harmony corrects, while cosine-similarity recall depends on absolute embedding geometry and benefits directly from batch alignment.


\newpage
\clearpage

\end{document}